\documentclass[runningheads]{llncs}

\usepackage{eccv}

\usepackage{eccvabbrv}

\usepackage{graphicx}
\usepackage{booktabs}
\usepackage{amsmath}
\usepackage{amssymb}
\usepackage{multirow}
\usepackage{wrapfig}
\usepackage{caption}
\usepackage[accsupp]{axessibility}  % Improves PDF readability for those with disabilities.

\usepackage{hyperref}

\usepackage{orcidlink}
\usepackage{marvosym}  % provides \Letter
\begin{document}

\title{Orthogonal Knowledge Refreshing for Domain-\\Incremental Object Detection} 

\titlerunning{Orthogonal Knowledge Refreshing}

\author{Aoting Zhang\inst{1,3} \and Dongbao Yang\inst{2}\textsuperscript{\ensuremath{\dagger}} \and Chang Liu\inst{4} \and Xiaopeng Hong\inst{5} \and  \\ Can Ma\inst{1,3} \and Yu Zhou\inst{2}\textsuperscript{\ensuremath{\dagger}}}
% \author{Aoting Zhang\inst{1,3}\orcidlink{0000-0003-2173-6327} \and Dongbao Yang\inst{2}\orcidlink{0000-0001-8628-411X} \and Chang Liu\inst{4}\orcidlink{0000-0001-6747-0646} \and Xiaopeng Hong\inst{5}\orcidlink{0000-0002-0611-0636} \and Can Ma\inst{1}\orcidlink{0009-0004-2307-5002} \and Yu Zhou\inst{2}\orcidlink{0000-0003-4188-9953}}
% Second Author\inst{2,3}\orcidlink{1111-2222-3333-4444} \and
% Third Author\inst{3}\orcidlink{2222--3333-4444-5555}}

\authorrunning{A.~Zhang et al.}

\institute{Institute of Information Engineering, Chinese Academy of Sciences \and
Nankai University \and
School of Cyber Security, University of Chinese Academy of Sciences \and 
Tsinghua University \and  Harbin Institute of Technology}
% \email{\{zhangaoting,macan\}@iie.ac.cn}\email{\{yangdongbao,yuzhou\}@nankai.edu.cn}}

\maketitle

\begingroup
\makeatletter
\renewcommand\@makefntext[1]{\noindent #1}
\makeatother
\renewcommand{\thefootnote}{}
\footnotetext{\textsuperscript{\ensuremath{\dagger}} Corresponding author.}
\endgroup

\begin{abstract}
Domain-incremental object detection (DIOD) requires models to continually adapt to new domains while preserving prior knowledge. Recently, parameter-efficient fine-tuning offers a promising avenue, wherein a pre-trained model is frozen and a small number of learnable parameters are injected for downstream tasks. However, these methods risk overwriting critical past knowledge, triggering inter-domain interference and performance degradation. To address this challenge, we propose Orthogonal Knowledge Refreshing (OKR), a simple yet effective framework for DIOD. OKR incrementally constructs independent domain-specific subspaces via dedicated low-rank branches for each domain, which are seamlessly fused for a holistic decision, enabling conflict-free capacity expansion without domain selection during inference. To minimize knowledge interference during fusion, we present a gradient-based orthogonal refreshing strategy that projects gradient updates of new domains onto the orthogonal complement of the fused historical subspace, supporting continual adaptation without forgetting. Moreover, to mitigate semantic fragmentation across domains, we enforce topology-aware consistency, aligning the semantic structures of old and new domains. Extensive experiments validate the superiority of OKR, outperforming the best exemplar-free method by significant margins of +5.6\% and +6.5\% mAP on the Pascal VOC and BDD100K series, respectively. 
% Code will be open-sourced upon acceptance.
\end{abstract}

\section{Introduction}
% \begin{figure}[t]
% 	\centering
% \includegraphics[width=\linewidth]{images/motivationv2.pdf}
% \caption{Comparison between OKR and existing PEFT-based approaches. (a) Previous methods select and optimize a subset of prompts from a prompt pool shared across all sessions, which overwrite prompts of old tasks, resulting in inter-domain interference. (b) OKR constructs domain-specific subspaces by expanding a LoRA branch for each new session while keeping old ones frozen, enabling updating the model without conflicts. Gradient orthogonalization-based update projects the gradient update in a direction orthogonal to the historical subspace, avoiding interference to old knowledge.}
% \label{fig: motivation}
% \end{figure}

\begin{figure*}
    \centering
    \includegraphics[width=\linewidth]{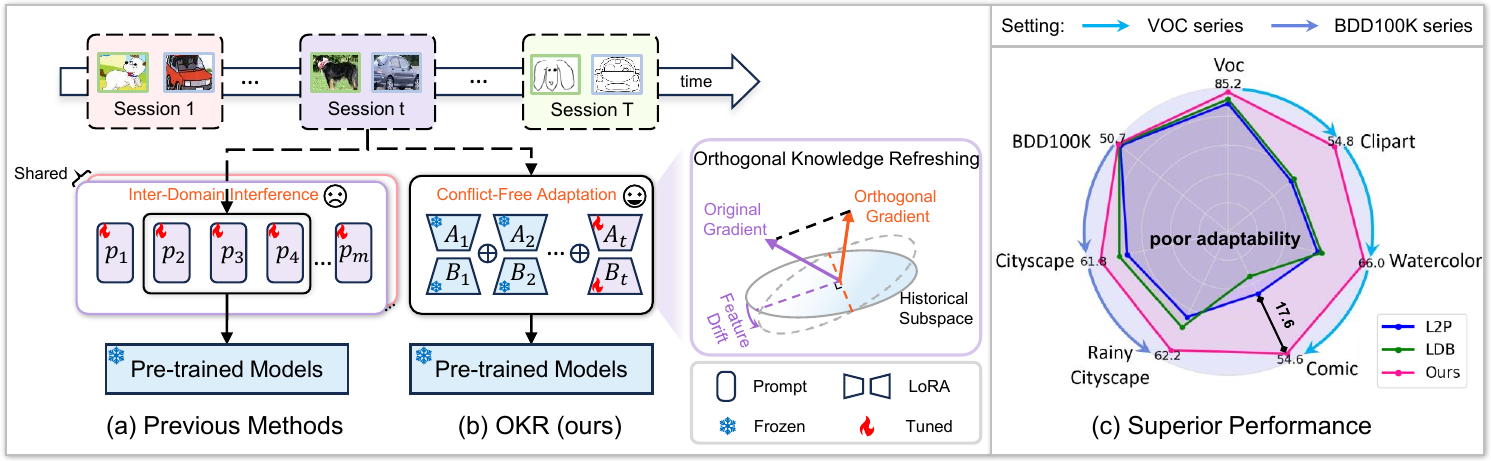}
    \vspace{-5mm}
    \caption{(a) Previous methods suffer from inter-domain interference due to shared parameter overwriting. (b) OKR enables conflict-free adaptation by expanding dedicated LoRA branches for each session. Through orthogonal knowledge refreshing, it projects gradients orthogonally to the fused historical subspace, preserving prior knowledge while bypassing domain routing. (c) OKR achieves superior performance, delivering a significant +17.6\% mAP gain on Comic dataset.}
    \label{fig:motivation}
    \vspace{-6mm}
\end{figure*}

Deep learning detectors have achieved notable success in object detection~\cite{ren2015faster, li2022exploring, yang2023pseudo, zhang2025class, cao2025devil}, largely due to robust representations learned from large-scale datasets. However, most frameworks assume identical training and testing distributions, leading to performance drops under domain shifts~\cite{yao2021multi}.
% and their performance drops when deployed in dynamic real-world scenarios with domain shifts~\cite{oza2023unsupervised, yao2021multi}. 
Domain adaptation partially addresses this by transferring knowledge from a labeled source to an unlabeled target domain, but it typically supports one-time adaptation and fails under sequential shifts. Real-world applications like autonomous driving demand continual adaptation to changing environments (e.g., from sunny to foggy to rainy) while retaining performance across all domains.
% In practical applications like autonomous driving, perception systems must adapt to continuously changing environments (e.g., from sunny to foggy to rainy) while retaining performance across all domains. 
This necessitates Domain-Incremental Object Detection (DIOD), a challenging yet underexplored setting requiring both adaptability and long-term knowledge retention.

The central challenge in DIOD is catastrophic forgetting~\cite{french1999catastrophic, zhang2026leveraging}, where adapting to new domains overwrites previously acquired knowledge. While rehearsal-based methods seek to alleviate this by periodically replaying historical exemplars~\cite{rolnick2019experience}, they are frequently constrained by stringent privacy protocols and memory overhead. Knowledge distillation methods~\cite{peng2021sid, yang2022rd, zhang2026focus} attempt to preserve prior knowledge by mimicking the output behavior of old models, yet they struggle to handle substantial domain shifts and are fundamentally limited by the static capacity of the underlying architecture. Furthermore, architecture-based paradigms~\cite{douillard2022dytox, zhang2025dca} expand task-specific sub-networks to accommodate new tasks, incurring computational costs and memory burdens that scale poorly with the number of tasks.

Recently, parameter-efficient fine-tuning (PEFT), particularly prompt learning~\cite{zhou2022conditional, Lu_2022_CVPR}, has emerged as a promising paradigm for domain incremental learning. For instance, L2P~\cite{wang2022learning} dynamically selects prompts from a shared pool, while S-Prompt~\cite{wang2022s} employs domain-specific prompts coupled with a K-NN-based routing mechanism for domain identification. Similarly, LDB~\cite{song2024non} learns domain biases with frozen base models and expands the predictors. However, these approaches overlook inter-domain interference, which undermines the trade-off between stability and plasticity. As illustrated in Figure~\ref{fig:motivation}(a), joint optimization within a shared parameter space inevitably distorts representations essential to prior domains, triggering detrimental feature drift. Moreover, prompt-based methods solely alter input embeddings, while bias-based ones apply additive shifts to neuron outputs, both exerting shallow control over the model's internal mechanisms~\cite{NEURIPS2023_eef6aecf}. This restricted modulation proves insufficient under significant domain shifts, often resulting in suboptimal adaptability and performance degradation (Figure~\ref{fig:motivation}(c)). By comparison, low-rank adaptation (LoRA) facilitates intrinsic modulation by directly updating weight matrices, enabling fine-grained control and superior flexibility for DIOD scenarios.

Motivated by these insights, we propose Orthogonal Knowledge Refreshing (OKR), a simple yet effective framework for DIOD. Specifically, OKR incrementally constructs an independent domain-specific subspace for each new domain by introducing a dedicated LoRA branch, enabling conflict-free capacity expansion. During adaptation, only the newly added LoRA is updated. Owing to LoRA’s inherent linearity, all branches are seamlessly fused for holistic prediction without redundant forward passes.
To minimize knowledge interference during fusion, we devise a gradient-based orthogonal refreshing strategy that constrains gradient updates of the new branch to be orthogonal to historical subspaces. Drawing on the connection between input features and gradients~\cite{saha2021gradient}, we approximate the composite prior subspaces via aggregated historical adaptation matrices and derive a projection matrix that redirects new gradients toward directions minimally correlated with existing knowledge, promoting stable adaptation without forgetting.
Furthermore, to mitigate semantic fragmentation caused by domain transitions, we impose topology-aware consistency to align semantic structures across domains, facilitating semantically coherent representations. As shown in Figure~\ref{fig:motivation}(c), OKR consistently outperforms existing methods across all domains, with particularly significant gains of +17.6\% mAP on the challenging Comic dataset, establishing new state-of-the-art performance.

Our contributions can be summarized as follows:
\begin{itemize}
    \item We propose orthogonal knowledge refreshing (OKR), an effective framework for DIOD that incrementally expands low-rank domain-specific subspaces, enabling conflict-free adaptation without past data or domain selection, while keeping parameters fixed via LoRA’s linear additivity.
    \item A gradient-based orthogonal refreshing strategy is designed to minimize interference with previously learned knowledge by constraining new-domain gradient directions to be orthogonal to historical subspaces.
    \item To alleviate semantic fragmentation across domains, we enforce topology-aware consistency, encouraging domain-specific subspaces to maintain semantically coherent and transferable structures.
    \item Extensive experiments on various settings demonstrate that OKR establishes new state-of-the-art performance in DIOD and effectively alleviates catastrophic forgetting.
\end{itemize}

\section {Related Works}

\subsection{Class Incremental Learning}
Class incremental learning aims to learn new classes sequentially over time without forgetting the previously learned classes~\cite{zhou2024class, zhang2025specifying}. Current mainstream methods for CIL can be divided into three categories. Replay-based methods mitigate catastrophic forgetting by reintroducing representative samples of historical data~\cite{bang2021rainbow, chaudhry2018efficient, yang2023one} to replay or synthesize pseudo samples with a generative model~\cite{gao2023ddgr} for training. Regularization-based methods impose additional constraints on network parameter updates, preventing drift~\cite{tang2021layerwise} in weights critical to previous tasks or leveraging knowledge distillation to bridge old and new models~\cite{yang2022multi, huang2024etag}. 
% Architecture-based methods dynamically increase the capacity of network~\cite{wang2022beef, wang2022learning}, retaining knowledge from previous tasks while expanding to learn new ones.
Architecture-based methods~\cite{wang2022beef, douillard2022dytox, wang2022learning} achieve strong performance among other competitors by allocating distinct subsets of parameters to various subtasks, facilitating the expansion of network architecture.
% initializing a new backbone for each task while retaining previous ones.

\subsection{Domain Incremental Learning}
In domain incremental learning, the label space remains fixed while domains vary sequentially~\cite{shi2023unified}. The goal is to adapt to new domains without retraining from scratch, while maintaining generalization on previously seen ones. Existing DIL methods can be categorized into replay-based~\cite{ding2023domain} and rehearsal-free approaches. We focus on the latter due to their greater practicality. CIFRCN~\cite{hao2019end} extends region proposal networks to support incremental domain adaptation while ERD~\cite{feng2022overcoming} adopts a response-based strategy. Among rehearsal-free methods, prompt-based tuning is particularly prominent. L2P\cite{wang2022learning} introduces a prompt pool and retrieves top-$k$ prompts to guide feature alignment in the pre-trained model. DualPrompt~\cite{wang2022dualprompt} further separates prompts into task-invariant G-Prompts and task-specific E-Prompts. S-Prompt~\cite{wang2022s} learns domain-specific prompts via K-means clustering and K-NN-based routing. LDB~\cite{song2024non} learns biases and predictors, while SOYO~\cite{Wang_2025_CVPR} enhances this by training a domain selector, both suffering from redundant computation due to repeated forward passes and limited cross-domain representational capacity. In this work, our orthogonal knowledge refreshing achieves conflict-free adaptation and efficient knowledge accumulation across incremental domains, bypassing the requirement for explicit domain routing.

\subsection{Source-Free Domain Adaptation}
A setting closely related to DIOD is source-free domain adaptation (SFDA), both aiming to ensure model robustness under evolving data distributions. SFDA adapts a source-trained model to an unlabeled target domain without access to the original source data. Existing SFDA approaches fall into three groups. The first constructs a pseudo source and aligns it with the target domain, treating the task as unsupervised domain adaptation. This is typically achieved via generative modeling~\cite{huang2021model, tian2021vdm} or target domain partitioning based on source hypotheses~\cite{du2024generation}. The second group extracts and transfers domain-invariant factors from the source to the target domain to align feature distributions~\cite{ding2022source}. 
The third strategy leverages auxiliary target-domain information, such as geometry or manifold structure~\cite{tang2021nearest}, beyond conventional pseudo-labeling~\cite{liang2020we}. However, SFDA is limited to one-time adaptation to a single target domain, whereas our work addresses continual adaptation across sequential domain shifts without forgetting prior domains.

\section{Preliminary}
\subsection{Observations}
\begin{figure}[t]
    \centering
    \includegraphics[width=\linewidth]{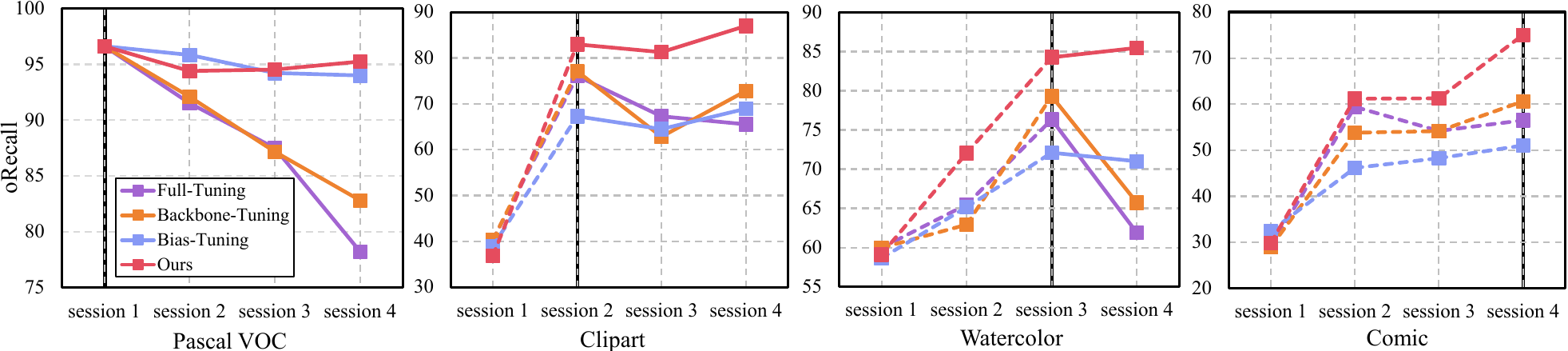}
    \vspace{-5mm}
    \caption{Class-agnostic object recall (oRecall) on different domains over learning sessions. Solid lines indicate the session in which each domain is learned.}
    \label{fig:placeholder}
    \vspace{-6mm}
\end{figure}

Performance degradation under domain shift primarily stems from feature distribution misalignment~\cite{wang_reducing, pmlr_v37_long15, xie2022active}. While mapping source and target features into an aligned representation space is effective for single-step adaptation~\cite{Saito_2019_CVPR, Li_sigma2023, VS_2021_CVPR}, this principle collapses in domain-incremental object detection. In DIOD, naively forcing sequential domains into a single space triggers a representation conflict, where updates for a new domain inevitably overwrite feature statistics vital to previous ones, leading to catastrophic forgetting. This suggests that effective DIOD requires more than simple alignment, but necessitates non-destructive feature accumulation: the ability to adapt to new distributions while isolating and preserving old-domain representations. 

To validate this premise, we conduct a diagnostic experiment measuring class-agnostic object recall (oRecall) across sequential domains. Unlike mAP, oRecall directly quantifies the model's ability to localize foreground objects independent of category-level classification, thus isolating the backbone’s representational plasticity. As observed in Figure~\ref{fig:placeholder}, backbone-tuning attains object recall comparable to full-tuning on new domains, confirming that the representational capacity of the backbone is sufficient to accommodate domain shifts. 
In contrast, bias-tuning which offers higher stability by updating fewer parameters, fails to capture complex distribution shifts, resulting in poor adaptation.
Critically, both full and backbone-tuning suffer from a sharp performance drop on non-current domains. This indicates that joint optimization within a shared parameter space drives forgetting by corrupting the previously learned objectness scoring and feature statistics.

These findings establish the core motivation for our orthogonal knowledge refreshing framework. Instead of shallow bias shifts or risky global updates, OKR performs non-interfering feature accumulation via parameter-efficient subspace learning. By training only 1.8\% of the parameters under explicit orthogonal constraints, we decouple the learning process into isolated subspaces, achieving a superior balance between rapid adaptation and long-term knowledge retention.

\subsection{Problem Formulation}
In DIOD, tasks involving $T$ distinct domains are presented sequentially, each accompanied by its own dataset $D\!=\!\left \{ D_{1},D_{2},\cdots,D_{T} \right \}$. For the $i$-th domain, the training set comprises $k$ image-annotation pairs: $D_i\!=\!\left \{(X_{i}^{1},Y_{i}^{1} ),\cdots,(X_{i}^{k},Y_{i}^{k} ) \right \}$, where $X^{k}_{i}$ denotes the input image and $Y^{k}_{i}$ represents its ground-truth annotation. A key challenge in DIOD lies in the evolving data distributions across domains while maintaining a consistent label space. Our approach follows a sequential training paradigm under the \textbf{exemplar-free} constraint. Specifically, at the $i$-th stage, the model is trained solely on $D_i$, without accessing historical data from $\left \{D_{1},\cdots,D_{i-1} \right\}$. The objective is to adapt to new domains while preserving performance on previously learned domains, ensuring knowledge retention across non-stationary data distributions. 

\section{Proposed Method}
\label{sec:method}

\begin{figure*}[t]
	\centering
\includegraphics[width=\textwidth]{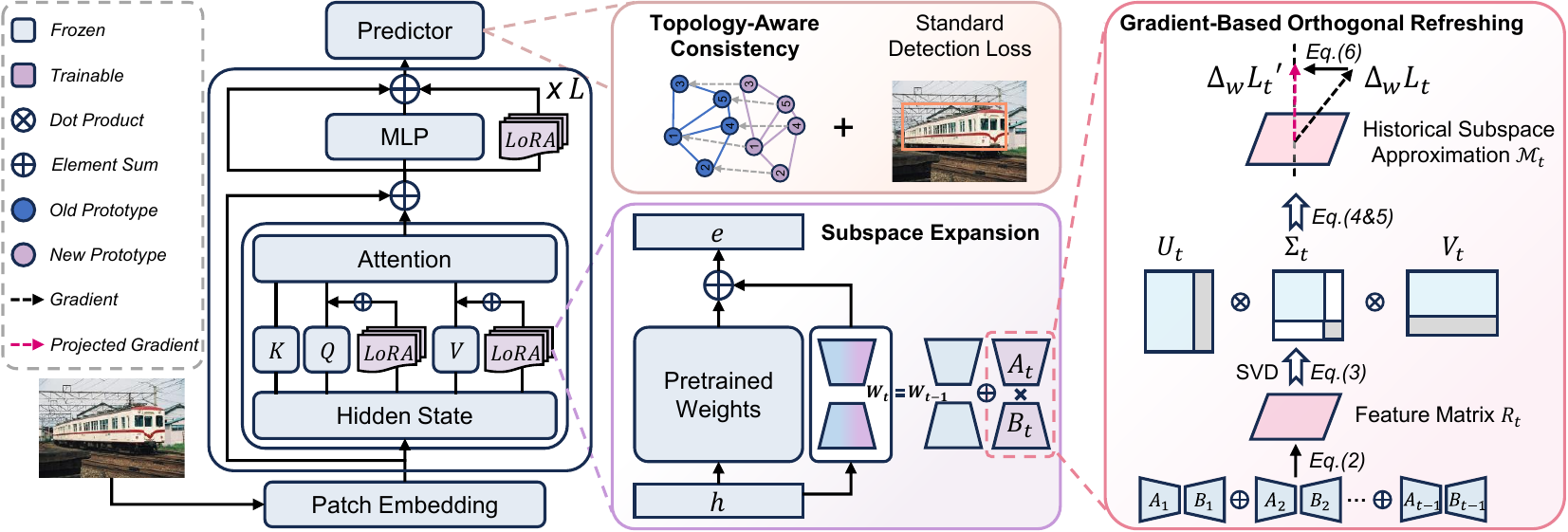}
\caption{Overview of OKR. To expand model capacity without conflicts, OKR incrementally constructs independent domain-specific subspaces by injecting dedicated LoRAs into the feature extractor. At session $t$, only the new LoRA (purple) is trainable, while previously ones (blue) remain frozen. During training and inference, all branches are seamlessly fused for holistic decision-making without redundant forward passes. To further suppress knowledge interference during weight fusion, we devise a gradient-based orthogonal refreshing strategy where $A_t$ and $B_t$ are updated by projecting the gradient $\nabla_{w} L_t$ onto the orthogonal complement of the approximated historical subspace $\mathcal{M}_t$, yielding the modified gradient $\nabla_{w} L_t'$. Moreover, we enforce topology-aware consistency to align class prototypes across domains, mitigating semantic fragmentation.}
\vspace{-4mm}
\label{fig: framework}
\end{figure*}
\subsection{Overall Structure} 
OKR is built upon the ViTDet~\cite{li2022exploring} detector, which is formulated as $h_{\Theta }(f_{\Phi }(\cdot ))$. Here, $f_{\Phi}(\cdot)$ denotes the transformer-based feature extractor with parameters $\Phi$, and $h_{\Theta}(\cdot)$ represents the linear predictor for object localization and classification with parameters $\Theta$. Following LDB~\cite{song2024non}, we initialize the feature extractor with ImageNet-1K pretrained weights $\Phi_{0}$ and train the base model $\Phi_{1}$ on the first domain $D_{1}$ under standard detection loss supervision. Subsequently, OKR employs PEFT for continuous domain adaptation in an exemplar-free manner eliminating the need for storing redundant exemplars. 

The architecture of our OKR is illustrated in Figure~\ref{fig: framework}. To enable conflict-free capacity expansion for accommodating multiple domains, we propose a low-rank subspace expansion mechanism that decouples the parameter space into domain-specific subspaces by incrementally injecting dedicated LoRAs into the feature extractor. At session $t$, only the LoRA parameters $A_t$ and $B_t$ are trainable. During both training and inference, the weights of all LoRAs are aggregated via LoRA's linear properties to support unified decision-making, avoiding redundant forward passes. To further suppress interference with historical knowledge, we introduce gradient-based orthogonal refreshing (GOR) which explicitly constrains the gradient updates of $A_t$ and $B_t$ to be orthogonal to the approximated historical subspaces $\mathcal{M}_t$. Finally, we integrate a topology-aware consistency (TAC) loss into the training objective, which aligns prototype structures across domains to mitigate semantic fragmentation and enhance representational coherence.

\subsection{Low-Rank Subspace Expansion}
% \textbf{Subspace Expansion with LoRA.} 
Owing to the inherent diversity and distinct characteristics of disparate domains, reusing shared LoRA parameter space across domains often induces interference and negative transfer of domain-specific knowledge. This challenge becomes more exacerbated in sequential domain-incremental scenarios, where adapting to a new domain may inadvertently distort representations optimized for previous domains. %To mitigate such conflicts, we decouple the parameter space into domain-specific subspaces by incrementally injecting dedicated LoRAs for new domains on top of a frozen backbone. Each LoRA is specialized for its domain, while historical branches remain frozen. This modular design enables incremental knowledge acquisition for new domains while preserving historical knowledge. During both training and inference, all LoRAs are seamlessly fused via weight aggregation, enabling unified prediction without redundant forward passes or explicit domain selection.
To alleviate such interference, we decouple the parameter space into domain-specific subspaces by incrementally refreshing dedicated LoRA components for new domains on top of a frozen backbone. Although each LoRA is optimized for its corresponding domain, the parameters are ultimately integrated through weight aggregation, maintaining a unified model with a \textit{fixed parameter budget}.  
As illustrated in Figure~\ref{fig: framework}, at session $t$, knowledge from the previous $t{-}1$ sessions is retained in frozen parameters $\left\{A_{1:t-1},B_{1:t-1}\right\}$, while only the newly $\left\{A_{t},B_{t}\right\}$ are trainable. The forward computation at session $t$ is:
\begin{equation}
    e = W_{0}h + \sum^{t}_{i=1}B_{i}A_{i}h = W_{t-1}h+B_{t}A_{t}h=W_{t}h, \label{eq: wab}
\end{equation}
where $W_{t}\!=\!W_{0}+\sum^{t}_{i=1}B_{i}A_{i}$ denotes the updated projection matrix (e.g., for query, value, or MLP layers), and $h$ is the input hidden state. For clarity, the layer index is omitted and this process is applied at each transformer block. 

This design facilitates incremental knowledge acquisition for new domains while preserving historical knowledge, without incurring parameter growth or redundant inference passes.
However, freezing previous branches alone is insufficient to eliminate knowledge interference. While weight fusion avoids branch switching and improves inference efficiency, it introduces implicit coupling. Gradient updates for the current domain can distort the latent feature space established by historical branches, causing feature drift and performance degradation on earlier domains. To maintain subspace independence without increasing parameter count, we enforce orthogonal gradient constraints during training, ensuring that updates for new domains occupy complementary directions in the shared low-rank space.

\subsection{Gradient-Based Orthogonal Refreshing}
Learning a new task exerts minimal interference on previously acquired knowledge when the gradient direction is orthogonal to the subspace spanned by features of old tasks~\cite{liang2023adaptive}. We explore the intrinsic relationship between gradient directions and input feature spaces, aiming to identify core gradient subspaces of old domains and enforce orthogonality during updates on new domains. Within the PEFT paradigm, LoRA can effectively capture low-rank gradient subspace~\cite{wang2023orthogonal}. Building upon this capability, we propose an effective gradient-based orthogonal refreshing strategy embedded in a low-rank subspace expansion framework for DIOD. Concretely, we construct a domain-specific LoRA module for each new domain and constrain its gradient updates to be orthogonal to the historical subspace. This design achieves dual objectives: preserving the low-level feature distributions of previous domains to prevent forgetting and enabling efficient, interference-free domain adaptation, a crucial aspect in existing research.

Before training on the $t$-th task, we approximate the gradient subspace $\mathcal{M}_t$ spanned by previous tasks. Let $\mathcal{M}_t^{\bot }$ denote the residual gradient subspace orthogonal to $\mathcal{M}_t$, where gradient updates along $\mathcal{M}_t$ cause high interference to old tasks, while those along $\mathcal{M}_{t}^{\bot}$ induce minimum or no interference. 
To suppress interference from the new task, we constrain its updates to lie within $\mathcal{M}_{t}^{\bot}$. Our objective is to identify the basis of $\mathcal{M}_{t}$ or $\mathcal{M}_{t}^{\bot}$ and ensure gradient steps are orthogonal to $\mathcal{M}_{t}$ during training.
Prior work~\cite{saha2021gradient} shows that the gradient update of a linear layer lies in the span of its inputs. Leveraging the linear superposition property of LoRA decompositions, we define the gradient space of task $t$ using its input matrix and the aggregated domain-specific parameters from previous tasks, denoted by $W_{t-1}=\sum_{i=1}^{t-1}B_{i}A_{i}$. Specifically, we feed training samples from $D_t$ into the model with linear layer parameterized by $W_{t-1}$ to obtain layer-wise feature matrix $R_{t}$, followed by Singular Value Decomposition (SVD):
\begin{align}
    &R_{t}=f_{\Phi}(W_{t-1}, D_t),\\
    &SVD(R_{t}) = U_{t}\Sigma_{t} (V_{t})^T,
\end{align}
where $R_{t}\in \mathbb{R}^{m\times n}$, $U_t\in \mathbb{R}^{m\times m}$ and $V_t\in \mathbb{R}^{n\times n}$ are orthogonal, and $\Sigma_t$ is a diagonal matrix with singular values sorted in descending order. To obtain a more compact representation, we apply a norm-based criterion to compute the $k$-rank approximation $(R_{t})_k$ of $R_t$ as follows: %\in\mathbb{R}^{m\times m}
\begin{equation}
    \left \| (R_t)_k \right \| _{F}^{2}\ge \epsilon \left \| R_t \right \|_{F}^{2} ,
\end{equation}
where $\left \| \cdot \right \|_{F}^{2}$ denotes the Frobenius norm, and $(R_t)_k$ retains the top-$k$ singular values along the diagonal. The threshold $\epsilon$ controls the number of selected singular values and basis vectors. A larger $\epsilon$ yields a higher-dimension feature space. The historical subspace is then spanned by the first $k$ columns of $U_t$, which can be presented as:
% $\mathcal{M}_{t}=span\left \{ u^1_t,u^2_t,\cdots,u^k_t \right \} $, .
\begin{equation}
    \mathcal{M}_{t}=span\left \{ u^1_t,u^2_t,\cdots,u^k_t \right \}.
\end{equation}

After approximating the gradient space, the new gradient is first projected onto $\mathcal{M}_t$, and the projected component is subtracted to ensure the remaining gradient lies in the orthogonal complement $\mathcal{M}_t^{\bot}$. Specifically, during training on task $t$, only new LoRA parameters $B_{t}$ and $A_{t}$ in each layer are updated. Let $\nabla_{w} L_t$ denote the gradient generated at each iteration, where $w\!=\!\left \{ A_t,B_t \right \} $. Then gradients are updated as follows:
\begin{equation}
    \nabla_{w} L_t = \nabla_{w} L_t - (\nabla_{w} L_t)\mathcal{M}_t(\mathcal{M}_t)^T.
\end{equation}

Beyond mitigating interference, our gradient refreshing strategy redirects learning toward residual subspaces beyond historical knowledge. This not only enhances the model's capacity to effectively acquire new knowledge but also establishes an optimal stability-plasticity balance.

\subsection{Topology-Aware Consistency}
While subspace expansion and orthogonal refreshing effectively capture cross-domain representations and alleviate inter-domain interference, the independently optimized domain-specific parameters introduce an inherent limitation. Isolated subspaces can fragment the global semantic coherence across domains. Although the label space remains shared, each subspace adapts solely to its respective domain, causing features of the same class to diverge in latent space. This disrupts intrinsic semantic alignment and gradually degrades cross-domain generalization. Such divergence accumulates with incremental steps, weakening the structural consistency of shared categories, which is crucial for robust multi-domain detection. To address this, we design topology-aware consistency, which explicitly regularizes semantic structures and bridges domain shifts.

We construct the topological structure in the feature space using class prototypes from the base domain, serving as a global reference for subsequent domains. Specifically, for each class $c$, we compute the prototype as: $\mu^c=\frac{1}{N^c}\sum_{i=1}^{N^c}p_i^c \cdot f_{\Phi}(x_i^c)$, where $f_{\Phi}$ is the feature extractor, $p_i^c$ is the classification confidence for object $i$, and $N^c$ is the number of instances in class $c$. The confidence-weighted average ensures robust prototype representation. The resulting prototype set $\mu_1=[\mu_1^1,\cdots,\mu_1^N]$ reflects the fundamental semantic topology and remains invariant throughout incremental training process. For a new domain $D_t$, we obtain updated prototypes $\mu_t=[\mu_t^1,\cdots,\mu_t^N]$ and encourage them to align with their corresponding base prototypes with the same class, thereby mitigating fragmentation caused by parameter-isolated subspace learning:
\begin{equation}
    L_{tac}=\frac{1}{N} \sum_{i=1}^{N} \sum_{j=1}^{N}w_t^{ij}\cdot (1-cos(\mu^{i}_{t},\mu^{j}_{1})), 2\le t\le T,
\end{equation}
where $w_{ij}=\frac{exp(\mu_t^i\cdot\mu_1^j)}{\sum_{k=1}^{N}exp(\mu_t^i\cdot\mu_1^k)} $ denotes the matching confidence between new and old prototypes. Higher semantic similarity yields greater alignment weights to reduce intra-class divergence. $cos(\cdot,\cdot)$ represents the cosine similarity function. Minimizing $L_{tac}$ promotes closer alignment between each new prototype and its corresponding base prototype, thereby enhancing intra-class compactness and maintaining structural consistency across domains. 

\subsection{Optimization}
Combining the detection loss $L_{det}$ and topology-aware consistency $L_{tac}$, the final objective when learning the new task $t$ ($2\le t \le T$) can be expressed as follows:
\begin{equation}
    L_{final} = L_{det} + L_{tac},
\end{equation}
Note that when learning the base domain ($t=1$), the model is optimized solely using the detection loss.
\section{Experiments}

\subsection{Experimental Setups}
\noindent\textbf{Datasets and Evaluation Metrics.} We conduct extensive experiments on three DIOD benchmarks with a broad spectrum of domain shifts, including style, scene and corruption variation. The \textit{Pascal VOC series} comprises four domains with diverse styles: VOC 2007~\cite{everingham2010pascal}, Clipart, Watercolor and Comic~\cite{inoue2018cross}, with six shared object categories across all datasets. The \textit{BDD100K series} focuses on scene-level shifts in autonomous driving, including BDD100k~\cite{yu2020bdd100k}, Cityscape~\cite{cordts2016cityscapes} and Rainy Cityscape~\cite{hu2019depth}. BDD100k is among the largest and most diverse driving datasets publicly available. The \textit{VOC-Corruption series} includes 15 domains with low-level visual corruptions~\cite{hendrycks2019benchmarking} applied to VOC images, serving as a fine-grained benchmark for robustness under distributional perturbations. We adopt the mean average precision (mAP) at a 0.5 IoU threshold as the evaluation metric. At each session $t$, after training on data $X^t$, we evaluate the model on the test sets of all previously seen domains.

\textbf{Implementation Details.}
We follow the implementation setup of previous work~\cite{song2024non} to ensure consistency and comparability. Our framework is built on ViTDet, using ViT-Base as the backbone, initialized with ImageNet-1K pretrained weights. All experiments are conducted under a strict rehearsal-free setting and performed using four NVIDIA RTX 3090 GPUs with a total batch size of 8. We adopt the AdamW optimizer with a learning rate of 2e-4 and a weight decay of 0.1. The LoRA module is configured with a rank of 16. After training on the first domain, all parameters are frozen except for a small subset introduced during parameter-efficient fine-tuning.

\begin{table*}[t]
\caption{Comparison results on Pascal VOC and BDD100K series. The best and second exemplar-free methods are in \textbf{bold} and \underline{underline}.}
\vspace{-8mm}
\label{tab: voc-bdd-series}
    \begin{center}
\resizebox{\textwidth}{!}{
        \begin{tabular}{l|c|c|cccc|c|ccc}
        \toprule
        \multirow{2}{*}{Methods} & \multirow{2}{*}{Source} & \multicolumn{5}{c|}{Pascal VOC series} & \multicolumn{4}{c}{BDD100K series} \\ \cmidrule{3-11} & & Exemplar & Session 2 & Session 3 & Session 4 &  $\Delta$mAP & Exemplar & Session 2 & Session 3 & $\Delta$mAP \\
            \midrule
             Upper-bound & - & - & 72.6 & 69.4 & 67.7 &- & -& 58.7 &59.1 & - \\ \midrule
             TP-DIOD-B~\cite{ding2023domain} & TCSVT 23 & 150/domain &65.8& 62.1 &{57.5} & -7.7 &200/domain & {53.4} & {51.5} & -6.7 \\ \midrule
             FT-Seq& - & \multirow{12}{*}{0/domain} &57.5            & 52.6            & 49.5 & -15.7 & \multirow{12}{*}{0/domain} &51.6 &43.6 &-14.6 \\
             FT-FC & - & &59.1            & 54.4            & 44.2& -21.0 & & 51.3 &48.0 & -10.2\\
             MCC~\cite{jin2020minimum} & ECCV 20 & &  47.6            & 34.4            & 23.7& -41.5 & & 44.3 &36.1&-22.1\\
             IRG~\cite{vs2023instance} & CVPR 23 && 51.5            & 43.7            & 33.2& -32.0 & &49.3 &38.7&-19.5\\
             LwF~\cite{li2017learning} & TPAMI 17& & 60.4            & 53.6            & 53.2& -12.0 & &52.1 &44.1&-14.1\\
             PASS~\cite{zhu2021prototype} & CVPR 21& & 61.7            & 51.4            & 49.8&  -12.7 & &51.7& 43.3&-14.9\\
             L2P~\cite{wang2022learning} & CVPR 22 & & 59.9            & 55.2            & 45.5& -19.7 & & 51.5 &47.7&-10.5\\
             S-Prompt~\cite{wang2022s} & NeurIPS 22&  & 59.4            & 54.3            & 45.0& -20.2 & &51.6& 49.4&-8.8\\
             CIFRCN~\cite{hao2019end} & ICME 19 & & 65.3            & 57.7            & 53.5& -11.7 & &51.8 &48.9&-9.3\\
            ERD~\cite{feng2022overcoming} &CVPR 22 & & 58.9            & 50.7            & 48.7& -16.5 &  &51.1 &48.1& -10.1 \\
            LDB~\cite{song2024non} & AAAI 25 & & {68.1}            & {64.2}            & {56.8} & -8.4 & & {52.3} & {51.1}& -7.1 \\ 
            LDB+SOYO~\cite{Wang_2025_CVPR} & CVPR 25 & & \underline{69.2} & \underline{65.3} & \underline{59.6} & -5.6 & & \underline{52.4} & \underline{51.7} & -6.5 \\ \midrule
            OKR (ours) & - & 0/domain  & \textbf{73.0} & \textbf{67.3} & \textbf{65.2} & -0.0 & 0/domain  & \textbf{57.6}& \textbf{58.2} & -0.0 \\
            
            \bottomrule
        \end{tabular}}
    \end{center}
\vspace{-5mm}
\end{table*}

\begin{table}[t]
\caption{Comparison with other PEFT-based methods at the final session on VOC series and BDD100K series.}
\vspace{-7mm}
\label{tab: voc-bdd-series-each}
\small
\setlength{\tabcolsep}{1.5mm}
    \begin{center}
\resizebox{\linewidth}{!}{
        \begin{tabular}{l|ccccc|cccc}
        \toprule
            % \multirow{1}{*}{Methods}& \multicolumn{5}{c}{Pascal VOC Series}  \\ \cmidrule{2-6}
            Method& VOC & Clipart & Watercolor & \multicolumn{1}{c|}{Comic} & mAP & BDD100K & Cityscape & \multicolumn{1}{c|}{Rainy} & {mAP} \\
            \midrule
             L2P~\cite{wang2022learning} & 78.2 & 32.5 & 43.3 & \multicolumn{1}{c|}{27.8} & 45.6 & 49.7& 48.8 &\multicolumn{1}{c|}{44.8} & {47.7} \\
             S-Prompt~\cite{wang2022s} & 80.8 &33.9 &45.2 &\multicolumn{1}{c|}{20.1}& 45.0  & \underline{51.6} &52.0 &\multicolumn{1}{c|}{44.7}& {49.4} \\
             LDB~\cite{song2024non} & \underline{82.4} & \underline{50.1} & \underline{57.5} & \multicolumn{1}{c|}{\underline{37.0} }& \underline{56.8} &  50.3 & \underline{52.7} & \multicolumn{1}{c|}{\underline{50.2}} & {\underline{51.1}} \\
             \midrule
            OKR (ours) & \textbf{85.2}	& \textbf{54.8}	& \textbf{66.0}	& \multicolumn{1}{c|}{\textbf{54.6}} & \textbf{65.2} & \textbf{50.7}	& \textbf{61.8}	& \multicolumn{1}{c|}{\textbf{62.2}} & {\textbf{58.2}} \\
            \bottomrule
        \end{tabular}}
    \end{center}
    \vspace{-7mm}
\end{table}

\subsection{Comparison Results}
\textbf{Results on Pascal VOC Series.} The left section of Table~\ref{tab: voc-bdd-series} presents the experimental results on Pascal VOC series, where the model undergoes sequential adaptation across four domains: Pascal VOC 2007 $\to$ Clipart $\to$ Watercolor $\to$ Comic, evaluating the model's ability to retain prior knowledge and address escalating visual style variations. OKR consistently outperforms the best exemplar-free methods LDB and LDB+SOYO, achieving significant gains of +8.4\% and +5.6\% mAP in the final session. Notably, even when compared with the exemplar-based method TP-DIOD-B~\cite{ding2023domain}, our method demonstrates superior generalization by exceeding its performance by +7.7\% at Session 4. Furthermore, OKR exhibits significant advantages over recent prompt-based and SFDA paradigms. For instance, at Session 4, it surpasses IRG~\cite{vs2023instance} by +32.0\% and L2P by +19.7\%, underscoring the effectiveness of our orthogonal knowledge refreshing strategy in adapting to new domain distributions.

\begin{figure}[t]
\centering

\begin{minipage}[t]{0.49\linewidth}
\vspace{0pt}
    \centering
    \includegraphics[width=\linewidth]{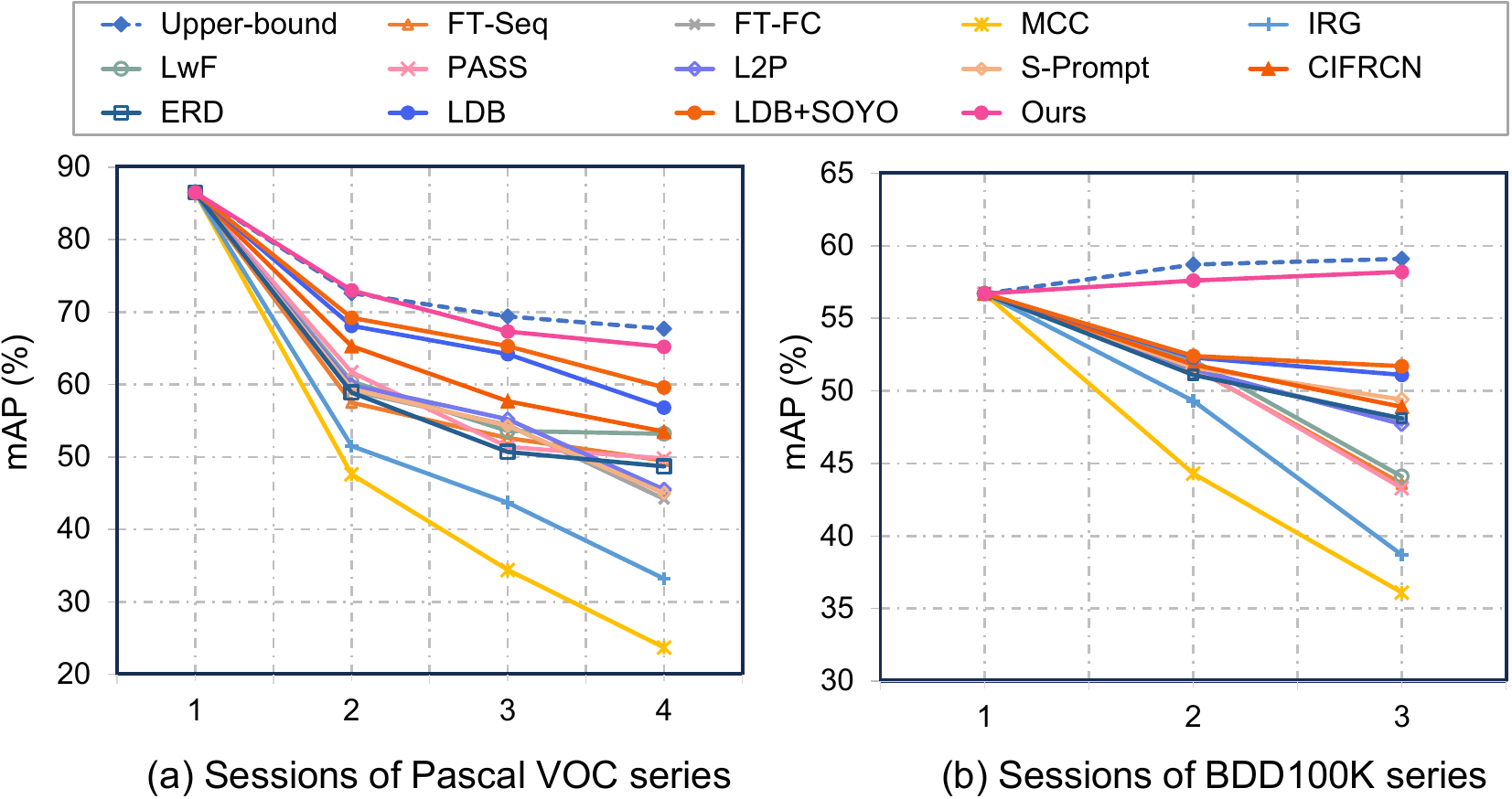}
    \vspace{-6mm}
    \captionsetup{type=figure}
    \caption{Performance across sessions on VOC and BDD100K series. OKR consistently outperforms PEFT-based methods.}
    \label{fig: voc-bdd-curve}
\end{minipage}
\hfill
\begin{minipage}[t]{0.49\linewidth}
\vspace{0pt}
\vspace{-3mm}
    \centering
    \captionsetup{type=table}
    \caption{Performance evolution on VOC-Corruption series.}
    \label{tab: voc-c-series}
    \vspace{2mm}

    \scriptsize
    \setlength{\tabcolsep}{0.6mm}
    \resizebox{\linewidth}{!}{
    \begin{tabular}{l|cccc}
    \toprule
    \multirow{2}{*}{Methods}  & \multicolumn{4}{c}{Average precision in each session (\%)} \\
    \cmidrule{2-5}
    & Session 2 & Session 3 & Session 4 & Session 5\\ 
    \midrule
    Upper-bound & 65.8 & 63.3 & 78.8 & 74.3 \\
    \midrule
    FT-Seq & \textbf{70.5} & 49.7 & \underline{55.3} & \underline{52.7} \\
    FT-FC  & 32.3 & 28.6 & 38.3 & 35.2\\
    L2P~\cite{wang2022learning} & 40.2 & 28.4 & 37.5 & 32.3\\
    S-Prompt~\cite{wang2022s} & 39.5 & 27.9 & 36.2 & 32.6\\
    ERD~\cite{feng2022overcoming} & 38.9 & 23.8 & 31.3 & 30.3 \\
    LDB~\cite{song2024non} & 52.3 & \underline{49.8} & 54.8 & 50.7\\
    \midrule
    OKR (ours) & \underline{67.3} & \textbf{54.6} & \textbf{62.9} & \textbf{61.0} \\
    \bottomrule
    \end{tabular}}
\end{minipage}

\vspace{-2mm}
\end{figure}

\textbf{Results on BDD100K Series.}
As displayed in the right section of Table~\ref{tab: voc-bdd-series}, OKR achieves consistent improvements on the BDD100K series, where domains progress in a realistic autonomous driving sequence: BDD100K $\to$ Cityscape $\to$ Rainy Cityscape. OKR surpasses all non-exemplar-based baselines, including LDB (+5.3\% and +7.1\%) and LDB+SOYO (+5.2\% and +6.5\%). Figure~\ref{fig: voc-bdd-curve} visualizes the curves of average precision with the session number increasing. We observe that OKR maintains a clear lead with performance margins steadily widening. Notably, our method approaches the upper-bound performance, highlighting OKR's exceptional resilience to domain shifts. 

Table~\ref{tab: voc-bdd-series-each} provides a more granular comparison against other parameter-efficient fine-tuning approaches in the final session. OKR a superior balance of knowledge retention and adaptability. Specifically, on the VOC series, it maintains high mAP scores on previously learned domains--85.2\% (VOC), 54.8\% (Clipart), and 66.0\% (Watercolor), while achieving 54.6\% on the newly introduced Comic domain. These results, echoed by similar trends on the BDD100K series, underscore OKR's enhanced cross-domain representational capacity and its unique ability to integrate knowledge across highly diverse visual contexts without explicit domain routing.

\begin{table*}[t]
% \caption{Detailed performance comparison after the final session across 16 domains in VOC-C series.}
\caption{Detailed performance comparison after the final session across 16 corruption domains in VOC-C series. Our method achieves superior balance between stability on previously learned domains and plasticity on new corruptions.}
\label{tab: voc-c-each}
\small
\vspace{-8mm}
\begin{center}
\resizebox{\textwidth}{!}{
\begin{tabular}{l|c|ccc|cccc|cccc|cccc|ccc}
\toprule
\multirow{2}{*}{Methods} & Org &\multicolumn{3}{c|}{Noise}                & \multicolumn{4}{c|}{Blur} & \multicolumn{4}{c|}{Weather} & \multicolumn{4}{c|}{Digital} &  \multirow{2}{*}{Avg.} & \multirow{2}{*}{Old} & \multirow{2}{*}{New}\\ \cmidrule{2-17}
& {Clean}& Gau &Sht& Imp& Def& Gls& Mtn& Zm &Snw& Frs &Fog &Brt& Cnt& Els &Px &Jpg  & \\
\midrule 
% Upper-bound & 64.2&67.2&66.1&64.5&65.7&67.5&63.6&77.3&76.3&78.1&83.8&67.3&79.4&76.1&74.4& 87.0 & 72.4 & 71.8&74.3 \\ \midrule
% FT-Seq  & 32.1 & 34.7 & 37.0 & 38.9 & 42.7 & 35.9 & 31.0 & 60.4 & 59.7 & 57.9 & 70.0 & \textbf{62.8}  & \underline{69.0} & \underline{67.1} & \underline{66.5} & 76.8 & \underline{52.7} & 48.1 & \underline{66.4} \\ 
    
% FT-FC & 8.7& 13.8 & 11.8 & 18.4 & 16.3 & 24.9 & 26.2 & 47.6 & 55.6 & \underline{61.6} & 72.1 & 8.3 & 69.0 & 4.5 & 37.0 & \underline{87.4} & 35.2 & 37.0 & 29.7 \\
  
{L2P}~\cite{wang2022learning} &79.6& 13.4 & 18.1&13.5&16.1&18.3&19.0&23.1&36.8&37.2&49.3&45.8&33.1&37.9&37.2&{37.5} &32.3&30.9&36.4   \\
{S-Prompt}~\cite{wang2022s} &81.8& 16.1&22.4&17.1&20.9&23.8&25.1&30.9&38.6&38.4&53.8&39.9&28.4&29.2&25.7&29.6&32.6&34.1&28.2   \\
ERD~\cite{feng2022overcoming}  & 52.0& 13.6 & 18.3 & 13.7 & 15.8 & 18.0 & 18.7 & 22.7 & 33.8 & 34.2 & 45.3 & 42.1 & 35.4 & 40.6 & \underline{39.8} & \underline{40.1}  &30.3 & 27.4 &39.0
  \\
    LDB~\cite{song2024non} & \underline{83.9}& \underline{41.3} & \underline{43.3} & \underline{43.1} & \underline{43.3} & \textbf{45.9} & \textbf{46.5} & \underline{37.3} & \underline{62.8} &	 \underline{65.0} & \underline{58.7} & \underline{75.7} & \underline{39.5} & \underline{64.5} & 23.5 & 37.4  & \underline{50.7} & \underline{53.9} & \underline{41.2}
 \\ \midrule
 OKR (ours)  & \textbf{87.9}& \textbf{49.4} & \textbf{55.0} & \textbf{53.4} & \textbf{43.9}	& \underline{44.2} & \underline{45.9} & \textbf{39.9} & \textbf{69.4} & \textbf{68.5} & \textbf{72.3} & \textbf{81.3} & \textbf{51.2}	& \textbf{77.8} & \textbf{68.8} & \textbf{66.8}  & \textbf{61.0} & \textbf{59.3} & \textbf{66.5}
\\ 
\bottomrule
\end{tabular}}
\vspace{-8mm}
\end{center}
\end{table*}
\begin{table}[t]
 \caption{Ablation of components where FT-back means fine-tuning backbone, SE denotes low-rank Subspace Expansion, GOR is used for Gradient-based Orthogonal Refreshing, and TAC for Topology-Aware Consistency.}\label{tab: ablation-component}
% \small
\vspace{-6mm}
\setlength{\tabcolsep}{2.5mm}
    \begin{center}
    \resizebox{0.8\textwidth}{!}{
        \begin{tabular}{c|cccc|ccc}
        \toprule
           Exp.& FT-Back &  SE & GOR & TAC & Session 2 & Session 3  & Session 4 \\ \midrule
             \#1&-& - & - & -& 59.1 &54.4 &44.2 \\ 
             \#2&$\checkmark$ & - & - & -& 67.7 & 57.9 & 49.5  \\
             \#3&$\checkmark$ & $\checkmark$ & - & - & 72.4 & 64.4 & 60.9
 \\
             \#4&$\checkmark$ & $\checkmark$ & $\checkmark$ & - & 72.4&66.9&64.9  \\
             \#5&$\checkmark$ & $\checkmark$ & - & $\checkmark$ & 72.6 & 66.0 & 63.3  \\
             
             \#6&$\checkmark$ & -&- & $\checkmark$ & 67.7 & 59.8      & 60.2   \\
             \#7&$\checkmark$ & $\checkmark$ & $\checkmark$ & $\checkmark$ & \textbf{73.0} & \textbf{67.3}& \textbf{65.2} 
\\ \bottomrule
\end{tabular}}
\vspace{-6mm}
    \end{center}
\end{table}

\textbf{Results on VOC-Corruption Series.}
To assess the robustness under real-world conditions involving exposure to multiple domains within one session, we evaluate on a challenging 16-dataset sequence split into 5 incremental sessions based on corruption types: Clean $\to$ Noise (Gaussian, Shot, Impulse) $\to$ Blur (Defocus, Glass, Motion, Zoom) $\to$ Weather (Snow, Frost, Fog, Brightness) $\to$ Digital (Contrast, Elastic, Pixelate, Jpeg), simulating escalating environmental challenges from low-level noise to complex weather and digital distortions. Table~\ref{tab: voc-c-series} summarizes the results on all learned datasets after each session, while Table~\ref{tab: voc-c-each} reports the detailed performance across 16 domains after the final session, highlighting both stability (Old) and plasticity (New). LDB suffers from poor plasticity (41.2\% vs. 66.4\%), in which learning biases cannot overcome the inherent complexities of learning multiple domains concurrently. ERD experiences severe forgetting. OKR outperforms all baselines, achieving 61.0\% mAP in Session 5, +10.3\% higher than LDB. Delved into individual domains, our method maintains strong performance, striking a better stability-plasticity trade-off.
% under these compound shifts.

\subsection{Ablation Study}
{\textbf{Effect of Different Components.}}
We conduct extensive ablation studies on the Pascal VOC series to evaluate the individual contributions of each component in OKR, as summarized in Table~\ref{tab: ablation-component}. The baseline (Exp.\#1) trained without anti-forgetting constraints, suffers from catastrophic forgetting, reaching only 44.2\% mAP in the final session. In Exp.\#2, updating only the backbone parameters improves performance to 49.5\%, suggesting that the predictor is largely domain-agnostic and the primary challenge of domain shifts lies in feature-level adaptation. With this insight, our proposed subspace expansion (SE, Exp.\#3) enables conflict-free adaptation without overwriting previous parameters, yielding a +16.7\% gain in Session 4. Gradient-based orthogonal refreshing (GOR) in Exp.\#4 constrains gradient update to be orthogonal to past subspaces, acquiring unique domain-specific knowledge and preventing interference, which improves performance by +20.7\%. Separately, topology-aware consistency (TAC, Exp.\#6) independently boosts performance by 10.7\% by mitigating semantic fragmentation. The synergy between SE and TAC (Exp.\#5) confirms the necessity of structural alignment even across isolated subspaces. Finally, the complete OKR framework achieves the best result, establishing a powerful pipeline for DIOD.

\begin{figure}[t]
	\centering
\includegraphics[width=\linewidth]{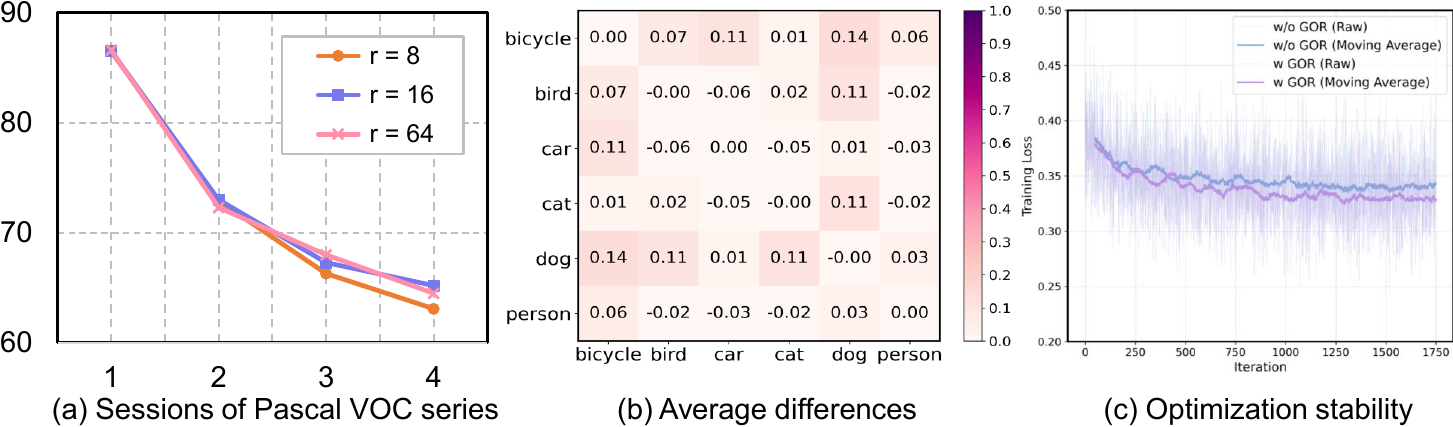}
\vspace{-6mm}
\caption{Further analysis of OKR. (a) OKR remains robust under different ranks of the expandable LoRA subspaces. (b) Topology-aware consistency preserves semantic topology across domain increments. (c) GOR maintains stable optimization, with similar training loss curves with and without orthogonal refreshing.}
\label{fig: ablation-rank}
    \vspace{-4mm}
\end{figure}

\textbf{Effect of Different Ranks.}
We investigate the impact of different ranks ($r\!=\!8,16,64$) for the expandable LoRA subspaces. As shown in Figure~\ref{fig: ablation-rank}(a), smaller ranks lead to a performance decline, due to insufficient capacity to represent complex domain-specific information. 
Interestingly, performance plateaus as the rank increases beyond 16, indicating that the model's gradient space possesses a relatively low intrinsic dimensionality. We set r=16 as it provides an optimal balance between model adaptability and computational efficiency.

\textbf{Analysis of Topology-Aware Consistency.}
To further validate the efficacy of $L_{tac}$, we examine the topological evolution of class prototypes during incremental learning sessions. Specifically, we construct semantic structures by calculating cosine similarity matrices between class prototypes within both base and target domains. The resulting aggregated difference matrix, averaged over all sessions, quantifies the cumulative structural drift throughout the learning process. As illustrated in Figure~\ref{fig: ablation-rank}(b), the values remain predominantly near zero, indicating that our framework preserves the underlying semantic topology with negligible distortion. This confirms that $L_{tac}$ effectively anchors emerging domain representations to the foundational semantic structure, successfully mitigating semantic fragmentation across sequential tasks.

\textbf{Effect of Orthogonality on Stability.}
Figure~\ref{fig: ablation-rank}(c) demonstrates that the proposed orthogonality constraint ensures stable and consistent optimization throughout the training process. The training loss curves with and without GOR exhibit nearly identical convergence, indicating that the orthogonal projection does not introduce additional optimization difficulties or instability. This stability aligns with the theoretical design of OKR, where GOR restricts new-domain updates to the residual gradient subspace orthogonal to historical ones. By isolating these updates, the model effectively minimizes interference with prior knowledge while maintaining subspace independence. Rather than disrupting the general convergence pattern, orthogonality enhances continual adaptation by directing the model to acquire complementary, domain-specific features, which accounts for the significant performance gains observed in the ablation study.

{\textbf{Computational Efficiency.}}
The proposed subspace expansion strategy introduces a marginal 1.8\% increase in total parameters, which is negligible relative to the substantial performance gains achieved. By leveraging the linear additivity of LoRA, all domain-specific branches are seamlessly merged into a unified parameter set after training, \textit{ensuring a fixed model size and eliminating the need for domain routing during inference}. Regarding training efficiency, the projection overhead is nearly negligible: SVD is performed only once per session, while per-step orthogonal refreshing involves solely low-rank matrix multiplications. This results in a mere 0.02\% increase in training time with only 1.8\% memory overhead. Empirically, OKR achieves an inference speed of 0.1348 s/sample with a 5.1 GB memory footprint on an RTX-3090, significantly outperforming LDB (0.2836 s/sample, 6.0 GB) due to OKR's efficient single-stage architecture.

\section{Conclusion}
In this work, we propose orthogonal knowledge refreshing (OKR), a simple yet effective framework for domain-incremental object detection. Grounded in the insight that backbone tuning provides sufficient plasticity but lacks stability due to representation overwriting, OKR leverages isolated domain-specific LoRA branches to facilitate conflict-free capacity expansion. By exploiting the linear additivity of low-rank matrices, all branches are seamlessly fused into a unified model, ensuring \textit{zero architectural overhead} and eliminating the need for complex domain routing during inference.
To eliminate knowledge interference during this fusion, we introduce a gradient-based orthogonal refreshing strategy that constrains new updates to the orthogonal complement of historical subspaces. This approach effectively redirects learning toward novel information while safeguarding the integrity of prior features. Moreover, we enforce topology-aware consistency to maintain semantic coherence across domains. Extensive experiments demonstrate OKR's superiority, establishing a new state-of-the-art of DIOD.

\section*{Acknowledgements}
This work is supported by the National Natural Science Foundation of China (Grant NO 62406318, 62376266, 62406167, U24B6012, 62376070 and 62076195).

% \clearpage\mbox{}Page \thepage\ of the manuscript.
% \clearpage\mbox{}Page \thepage\ of the manuscript.
% \clearpage\mbox{}Page \thepage\ of the manuscript.
% \clearpage\mbox{}Page \thepage\ of the manuscript.
% \clearpage\mbox{}Page \thepage\ of the manuscript. This is the last page.
% Now we have reached the maximum length of an ECCV \ECCVyear{} submission (excluding references and acknowledgements).
% References should start immediately after the main text, but can continue past p.\ 14 if needed. 
% \clearpage  % TODO FINAL: This \clearpage needs to be removed from both review and camera-ready versions.

\bibliographystyle{splncs04}
\bibliography{main}

@String(IJCV  = {Int. J. Comput. Vis.})

@String(CVPR  = {IEEE Conf. Comput. Vis. Pattern Recog.})

@String(ICCV  = {Int. Conf. Comput. Vis.})

@String(ECCV  = {Eur. Conf. Comput. Vis.})

@String(NeurIPS = {Adv. Neural Inform. Process. Syst.})

@String(ICML  = {Int. Conf. Mach. Learn.})

@String(ICLR  = {Int. Conf. Learn. Represent.})

@String(AAAI  = {AAAI})

@String(ICME  = {Int. Conf. Multimedia and Expo})

@String(TCSVT = {IEEE Trans. Circuit Syst. Video Technol.})

@String(PR    = {Pattern Recognition})

@String(IJCV  = {IJCV})

@String(CVPR  = {CVPR})

@String(ICCV  = {ICCV})

@String(ECCV  = {ECCV})

@String(NeurIPS = {NeurIPS})

@String(ICML  = {ICML})

@String(ICLR  = {ICLR})

@String(ICME  =	{ICME})

@String(TCSVT = {IEEE TCSVT})

@String(PR    = {PR})

@article{zhou2024class,
  title={Class-incremental learning: A survey},
  author={Zhou, Da-Wei and Wang, Qi-Wei and Qi, Zhi-Hong and Ye, Han-Jia and Zhan, De-Chuan and Liu, Ziwei},
  journal={IEEE TPAMI},
  year={2024},
}

@inproceedings{bang2021rainbow,
  title={Rainbow {M}emory: Continual learning with a memory of diverse samples},
  author={Bang, Jihwan and Kim, Heesu and Yoo, YoungJoon and Ha, Jung-Woo and Choi, Jonghyun},
  booktitle={CVPR},
  pages={8218--8227},
  year={2021}
}

@inproceedings{gao2023ddgr,
  title={{DDGR}: Continual learning with deep diffusion-based generative replay},
  author={Gao, Rui and Liu, Weiwei},
  booktitle={ICML},
  pages={10744--10763},
  year={2023}
}

@inproceedings{tang2021layerwise,
  title={Layerwise optimization by gradient decomposition for continual learning},
  author={Tang, Shixiang and Chen, Dapeng and Zhu, Jinguo and Yu, Shijie and Ouyang, Wanli},
  booktitle={CVPR},
  pages={9634--9643},
  year={2021}
}

@inproceedings{wang2022beef,
  title={{BEEF}: Bi-compatible class-incremental learning via energy-based expansion and fusion},
  author={Wang, Fu-Yun and Zhou, Da-Wei and Liu, Liu and Ye, Han-Jia and Bian, Yatao and Zhan, De-Chuan and Zhao, Peilin},
  booktitle={ICLR},
  year={2022}
}

@article{chaudhry2018efficient,
  title={Efficient lifelong learning with a-gem},
  author={Chaudhry, Arslan and Ranzato, Marc'Aurelio and Rohrbach, Marcus and Elhoseiny, Mohamed},
  journal={arXiv preprint arXiv:1812.00420},
  year={2018}
}

@inproceedings{yang2022multi,
  title={Multi-view correlation distillation for incremental object detection},
  author={Yang, Dongbao and Zhou, Yu and Zhang, Aoting and Sun, Xurui and Wu, Dayan and Wang, Weiping and Ye, Qixiang},
  booktitle={PR},
  volume={131},
  pages={108863},
  year={2022}
}

@inproceedings{huang2024etag,
  title={{eTag}: Class-incremental learning via embedding distillation and task-oriented generation},
  author={Huang, Libo and Zeng, Yan and Yang, Chuanguang and An, Zhulin and Diao, Boyu and Xu, Yongjun},
  booktitle={AAAI},
  volume={38},
  number={11},
  pages={12591--12599},
  year={2024}
}

@inproceedings{douillard2022dytox,
  title={Dytox: Transformers for continual learning with dynamic token expansion},
  author={Douillard, Arthur and Ram{\'e}, Alexandre and Couairon, Guillaume and Cord, Matthieu},
  booktitle={CVPR},
  pages={9285--9295},
  year={2022}
}

@inproceedings{wang2022learning,
	title={Learning to prompt for continual learning},
	author={Wang, Zifeng and Zhang, Zizhao and Lee, Chen-Yu and Zhang, Han and Sun, Ruoxi and Ren, Xiaoqi and Su, Guolong and Perot, Vincent and Dy, Jennifer and Pfister, Tomas},
	booktitle={CVPR},
	pages={139--149},
	year={2022}
}

@inproceedings{shi2023unified,
  title={A unified approach to domain incremental learning with memory: Theory and algorithm},
  author={Shi, Haizhou and Wang, Hao},
  booktitle={NeurIPS},
  volume={36},
  pages={15027--15059},
  year={2023}
}

@InProceedings{wang2022s,
  title={{S-Prompts} learning with pre-trained transformers: An occam’s razor for domain incremental learning},
  author={Wang, Yabin and Huang, Zhiwu and Hong, Xiaopeng},
  booktitle = {NeurIPS},
  volume={35},
  pages={5682--5695},
  year={2022}
}

@inproceedings{song2024non,
  title={Non-exemplar domain incremental object detection via learning domain bias},
  author={Song, Xiang and He, Yuhang and Dong, Songlin and Gong, Yihong},
  booktitle={AAAI},
  volume={38},
  number={13},
  pages={15056--15065},
  year={2024}
}

@inproceedings{huang2021model,
  title={{Model Adaptation}: Historical contrastive learning for unsupervised domain adaptation without source data},
  author={Huang, Jiaxing and Guan, Dayan and Xiao, Aoran and Lu, Shijian},
  booktitle={NeurIPS},
  volume={34},
  pages={3635--3649},
  year={2021}
}

@inproceedings{tian2021vdm,
  title={{VDM-DA}: Virtual domain modeling for source data-free domain adaptation},
  author={Tian, Jiayi and Zhang, Jing and Li, Wen and Xu, Dong},
  booktitle={IEEE TCSVT},
  volume={32},
  number={6},
  pages={3749--3760},
  year={2021}
}

@inproceedings{du2024generation,
  title={Generation, augmentation, and alignment: A pseudo-source domain based method for source-free domain adaptation},
  author={Du, Yuntao and Yang, Haiyang and Chen, Mingcai and Luo, Hongtao and Jiang, Juan and Xin, Yi and Wang, Chongjun},
  booktitle={ML},
  volume={113},
  number={6},
  pages={3611--3631},
  year={2024}
}

@inproceedings{ding2022source,
  title={Source-free domain adaptation via distribution estimation},
  author={Ding, Ning and Xu, Yixing and Tang, Yehui and Xu, Chao and Wang, Yunhe and Tao, Dacheng},
  booktitle={CVPR},
  pages={7212--7222},
  year={2022}
}

@inproceedings{liang2020we,
  title={Do we really need to access the source data? source hypothesis transfer for unsupervised domain adaptation},
  author={Liang, Jian and Hu, Dapeng and Feng, Jiashi},
  booktitle={ICML},
  pages={6028--6039},
  year={2020}
}

@article{tang2021nearest,
  title={Nearest neighborhood-based deep clustering for source data-absent unsupervised domain adaptation},
  author={Tang, Song and Yang, Yan and Ma, Zhiyuan and Hendrich, Norman and Zeng, Fanyu and Ge, Shuzhi Sam and Zhang, Changshui and Zhang, Jianwei},
  journal={arXiv preprint arXiv:2107.12585},
  year={2021}
}

@inproceedings{ding2023domain,
  title={Domain incremental object detection based on feature space topology preserving strategy},
  author={Ding, Li and Song, Xiang and He, Yuhang and Wang, Changxin and Dong, Songlin and Wei, Xing and Gong, Yihong},
  booktitle={IEEE TCSVT},
  volume={34},
  number={1},
  pages={424--437},
  year={2023}
}

@inproceedings{jin2020minimum,
  title={Minimum class confusion for versatile domain adaptation},
  author={Jin, Ying and Wang, Ximei and Long, Mingsheng and Wang, Jianmin},
  booktitle={ECCV},
  pages={464--480},
  year={2020},
}

@inproceedings{vs2023instance,
  title={Instance relation graph guided source-free domain adaptive object detection},
  author={VS, Vibashan and Oza, Poojan and Patel, Vishal M},
  booktitle={CVPR},
  pages={3520--3530},
  year={2023}
}

@inproceedings{zhu2021prototype,
  title={Prototype augmentation and self-supervision for incremental learning},
  author={Zhu, Fei and Zhang, Xu-Yao and Wang, Chuang and Yin, Fei and Liu, Cheng-Lin},
  booktitle={CVPR},
  pages={5871--5880},
  year={2021}
}

@inproceedings{hao2019end,
  title={An end-to-end architecture for class-incremental object detection with knowledge distillation},
  author={Hao, Yu and Fu, Yanwei and Jiang, Yu-Gang and Tian, Qi},
  booktitle={ICME},
  pages={1--6},
  year={2019}
}

@inproceedings{feng2022overcoming,
  title={Overcoming catastrophic forgetting in incremental object detection via elastic response distillation},
  author={Feng, Tao and Wang, Mang and Yuan, Hangjie},
  booktitle={CVPR},
  pages={9427--9436},
  year={2022}
}

@inproceedings{li2017learning,
  title={Learning without forgetting},
  author={Li, Zhizhong and Hoiem, Derek},
  booktitle={IEEE TPAMI},
  volume={40},
  number={12},
  pages={2935--2947},
  year={2017}
}

@inproceedings{ren2015faster,
  title={Faster {R-CNN}: Towards real-time object detection with region proposal networks},
  author={Ren, Shaoqing and He, Kaiming and Girshick, Ross and Sun, Jian},
  booktitle={NeurIPS},
  volume={28},
  pages={91--99},
  year={2015}
}

@inproceedings{li2022exploring,
  title={Exploring plain vision transformer backbones for object detection},
  author={Li, Yanghao and Mao, Hanzi and Girshick, Ross and He, Kaiming},
  booktitle={ECCV},
  pages={280--296},
  year={2022},
}

@inproceedings{yao2021multi,
  title={Multi-source domain adaptation for object detection},
  author={Yao, Xingxu and Zhao, Sicheng and Xu, Pengfei and Yang, Jufeng},
  booktitle={ICCV},
  pages={3273--3282},
  year={2021}
}

@inproceedings{french1999catastrophic,
  title={Catastrophic forgetting in connectionist networks},
  author={French, Robert M},
  booktitle={TCS},
  volume={3},
  number={4},
  pages={128--135},
  year={1999},
}

@inproceedings{rolnick2019experience,
  title={Experience replay for continual learning},
  author={Rolnick, David and Ahuja, Arun and Schwarz, Jonathan and Lillicrap, Timothy and Wayne, Gregory},
  booktitle={NeurIPS},
  volume={32},
  pages={350--360},
  year={2019}
}

@inproceedings{peng2021sid,
  title={{SID}: Incremental learning for anchor-free object detection via Selective and Inter-related Distillation},
  author={Peng, Can and Zhao, Kun and Maksoud, Sam and Li, Meng and Lovell, Brian C},
  booktitle={CVIU},
  volume={210},
  pages={103229},
  year={2021}
}

@inproceedings{zhou2022conditional,
  title={Conditional prompt learning for vision-language models},
  author={Zhou, Kaiyang and Yang, Jingkang and Loy, Chen Change and Liu, Ziwei},
  booktitle={CVPR},
  pages={16816--16825},
  year={2022}
}

@article{saha2021gradient,
  title={Gradient projection memory for continual learning},
  author={Saha, Gobinda and Garg, Isha and Roy, Kaushik},
  journal={arXiv preprint arXiv:2103.09762},
  year={2021}
}

@inproceedings{liang2023adaptive,
  title={Adaptive plasticity improvement for continual learning},
  author={Liang, Yan-Shuo and Li, Wu-Jun},
  booktitle={CVPR},
  pages={7816--7825},
  year={2023}
}

@article{wang2023orthogonal,
  title={Orthogonal subspace learning for language model continual learning},
  author={Wang, Xiao and Chen, Tianze and Ge, Qiming and Xia, Han and Bao, Rong and Zheng, Rui and Zhang, Qi and Gui, Tao and Huang, Xuanjing},
  journal={arXiv preprint arXiv:2310.14152},
  year={2023}
}

@inproceedings{everingham2010pascal,
  title={The pascal visual object classes (voc) challenge},
  author={Everingham, Mark and Van Gool, Luc and Williams, Christopher KI and Winn, John and Zisserman, Andrew},
  booktitle={IJCV},
  volume={88},
  pages={303--338},
  year={2010},
}

@inproceedings{inoue2018cross,
  title={Cross-domain weakly-supervised object detection through progressive domain adaptation},
  author={Inoue, Naoto and Furuta, Ryosuke and Yamasaki, Toshihiko and Aizawa, Kiyoharu},
  booktitle={CVPR},
  pages={5001--5009},
  year={2018}
}

@inproceedings{yu2020bdd100k,
  title={Bdd100k: A diverse driving dataset for heterogeneous multitask learning},
  author={Yu, Fisher and Chen, Haofeng and Wang, Xin and Xian, Wenqi and Chen, Yingying and Liu, Fangchen and Madhavan, Vashisht and Darrell, Trevor},
  booktitle={CVPR},
  pages={2636--2645},
  year={2020}
}

@inproceedings{cordts2016cityscapes,
  title={The cityscapes dataset for semantic urban scene understanding},
  author={Cordts, Marius and Omran, Mohamed and Ramos, Sebastian and Rehfeld, Timo and Enzweiler, Markus and Benenson, Rodrigo and Franke, Uwe and Roth, Stefan and Schiele, Bernt},
  booktitle={CVPR},
  pages={3213--3223},
  year={2016}
}

@inproceedings{hu2019depth,
  title={Depth-attentional features for single-image rain removal},
  author={Hu, Xiaowei and Fu, Chi-Wing and Zhu, Lei and Heng, Pheng-Ann},
  booktitle={CVPR},
  pages={8022--8031},
  year={2019}
}

@article{hendrycks2019benchmarking,
  title={Benchmarking neural network robustness to common corruptions and perturbations},
  author={Hendrycks, Dan and Dietterich, Thomas},
  journal={arXiv preprint arXiv:1903.12261},
  year={2019}
}

@inproceedings{NEURIPS2023_eef6aecf,
 title = {Universality and Limitations of Prompt Tuning},
 author = {Wang, Yihan and Chauhan, Jatin and Wang, Wei and Hsieh, Cho-Jui},
 booktitle = {NeurIPS},
 pages = {75623--75643},
 volume = {36},
 year = {2023}
}

@inproceedings{wang2022dualprompt,
  title={{DualPrompt}: Complementary prompting for rehearsal-free continual learning},
  author={Wang, Zifeng and Zhang, Zizhao and Ebrahimi, Sayna and Sun, Ruoxi and Zhang, Han and Lee, Chen-Yu and Ren, Xiaoqi and Su, Guolong and Perot, Vincent and Dy, Jennifer and others},
  booktitle={ECCV},
  pages={631--648},
  year={2022}
}

@article{wang_reducing,
    title = {Reducing bi-level feature redundancy for unsupervised domain adaptation},
    journal = {PR},
    volume = {137},
    pages = {109319},
    year = {2023},
    author = {Mengzhu Wang and Shanshan Wang and Wei Wang and Li Shen and Xiang Zhang and Long Lan and Zhigang Luo},
}

@InProceedings{Saito_2019_CVPR,
    author = {Saito, Kuniaki and Ushiku, Yoshitaka and Harada, Tatsuya and Saenko, Kate},
    title = {Strong-Weak Distribution Alignment for Adaptive Object Detection},
    booktitle = {CVPR},
    month = {June},
    year = {2019}
}

@ARTICLE{Li_sigma2023,
  author={Li, Wuyang and Liu, Xinyu and Yuan, Yixuan},
  journal={IEEE TPAMI}, 
  title={{SIGMA++:} Improved Semantic-Complete Graph Matching for Domain Adaptive Object Detection}, 
  year={2023},
  volume={45},
  number={7},
  pages={9022-9040},
}

@InProceedings{VS_2021_CVPR,
    author    = {VS, Vibashan and Gupta, Vikram and Oza, Poojan and Sindagi, Vishwanath A. and Patel, Vishal M.},
    title     = {{MeGA-CDA}: Memory Guided Attention for Category-Aware Unsupervised Domain Adaptive Object Detection},
    booktitle = {CVPR},
    month     = {June},
    year      = {2021},
    pages     = {4516-4526}
}

@InProceedings{pmlr_v37_long15,
  title = 	 {Learning Transferable Features with Deep Adaptation Networks},
  author = 	 {Long, Mingsheng and Cao, Yue and Wang, Jianmin and Jordan, Michael},
  booktitle = 	 {ICML},
  pages = 	 {97--105},
  year = 	 {2015},
  volume = 	 {37},
  month = 	 {July},
}

@inproceedings{xie2022active,
  title={Active learning for domain adaptation: An energy-based approach},
  author={Xie, Binhui and Yuan, Longhui and Li, Shuang and Liu, Chi Harold and Cheng, Xinjing and Wang, Guoren},
  booktitle={AAAI},
  volume={36},
  number={8},
  pages={8708--8716},
  year={2022}
}

@InProceedings{Wang_2025_CVPR,
    author    = {Wang, Qiang and Song, Xiang and He, Yuhang and Han, Jizhou and Ding, Chenhao and Gao, Xinyuan and Gong, Yihong},
    title     = {Boosting Domain Incremental Learning: Selecting the Optimal Parameters is All You Need},
    booktitle = {CVPR},
    month     = {June},
    year      = {2025},
    pages     = {4839-4849}
}

@InProceedings{Lu_2022_CVPR,
    author    = {Lu, Yuning and Liu, Jianzhuang and Zhang, Yonggang and Liu, Yajing and Tian, Xinmei},
    title     = {Prompt Distribution Learning},
    booktitle = {CVPR},
    month     = {June},
    year      = {2022},
    pages     = {5206-5215}
}

@inproceedings{yang2023pseudo,
  title={Pseudo Object Replay and Mining for Incremental Object Detection},
  author={Yang, Dongbao and Zhou, Yu and Hong, Xiaopeng and Zhang, Aoting and Wei, Xin and Zeng, Linchengxi and Qiao, Zhi and Wang, Weipinng},
  booktitle={Proceedings of the 31st ACM International Conference on Multimedia},
  pages={153--162},
  year={2023}
}

@inproceedings{yang2023one,
  title={One-Shot Replay: Boosting Incremental Object Detection via Retrospecting One Object},
  author={Yang, Dongbao and Zhou, Yu and Hong, Xiaopeng and Zhang, Aoting and Wang, Weiping},
  booktitle={Proceedings of the AAAI Conference on Artificial Intelligence},
  volume={37},
  number={3},
  pages={3127--3135},
  year={2023}
}

@article{yang2022rd,
  title={{RD-IOD}: Two-level residual-distillation-based triple-network for incremental object detection},
  author={Yang, Dongbao and Zhou, Yu and Shi, Wei and Wu, Dayan and Wang, Weiping},
  journal={ACM Transactions on Multimedia Computing, Communications, and Applications},
  volume={18},
  number={1},
  pages={1--23},
  year={2022}
}

@inproceedings{zhang2025dca,
  title={{DCA}: Dividing and Conquering Amnesia in Incremental Object Detection},
  author={Zhang, Aoting and Yang, Dongbao and Liu, Chang and Hong, Xiaopeng and Shang, Miao and Zhou, Yu},
  booktitle={Proceedings of the AAAI Conference on Artificial Intelligence},
  volume={39},
  number={9},
  pages={9851--9859},
  year={2025}
}

@inproceedings{zhang2025specifying,
  title={Specifying what you know or not for multi-label class-incremental learning},
  author={Zhang, Aoting and Yang, Dongbao and Liu, Chang and Hong, Xiaopeng and Zhou, Yu},
  booktitle={Proceedings of the AAAI Conference on Artificial Intelligence},
  volume={39},
  number={21},
  pages={22345--22353},
  year={2025}
}

@article{zhang2026focus,
  title={{F}ocus, {A}lign, and {S}ustain: Counteracting Gradient Dilution in Incremental Object Detection},
  author={Zhang, Aoting and Yang, Dongbao and Liu, Chang and Hong, Xiaopeng and Zhou, Yu},
  journal={arXiv preprint arXiv:2606.15253},
  year={2026}
}

@article{zhang2026leveraging,
  title={Leveraging multi-modal and historical knowledge graphs for continual robot navigation},
  author={Zhang, Lin and Qian, Longyue and Li, Ruitong and Li, Teng and Zhang, Wei},
  journal={Visual Intelligence},
  volume={4},
  number={1},
  pages={15},
  year={2026}
}

@inproceedings{zhang2025class,
  title={Class-Agnostic Region-of-Interest Matching in Document Images},
  author={Zhang, Demin and Lyu, Jiahao and Shen, Zhijie and Zhou, Yu},
  booktitle={International Conference on Document Analysis and Recognition},
  pages={446--464},
  year={2025},
}

@article{cao2025devil,
  title={The devil is in fine-tuning and long-tailed problems: a new benchmark for scene text detection},
  author={Cao, Tianjiao and Lyu, Jiahao and Zeng, Weichao and Mu, Weimin and Zhou, Yu},
  journal={arXiv preprint arXiv:2505.15649},
  year={2025}
}
\end{document}